# Credibility Discounting in the Theory of Approximate Reasoning

Ronald R. Yager

Machine Intelligence Institute, Iona College

New Rochelle, NY 10801

## ABSTRACT

*We are concerned with the problem of introducing credibility type information into reasoning systems. The concept of credibility allows us to discount information provided by agents. An important characteristic of this kind of procedure is that a complete lack of credibility rather than resulting in the negation of the information provided results in the nullification of the information provided. We suggest a representational scheme for credibility qualification in the theory of approximate reasoning. We discuss the concept of relative credibility. By this idea we mean to indicate situations in which the credibility of a piece of evidence is determined by its compatibility with higher priority evidence. This situation leads to structures very much in the spirit of nonmonotonic reasoning.*

## 1. Introduction

In the construction of automated systems which aggregate and reason with pieces of evidence, an important consideration is the amount of credibility associated with the individual pieces of evidences. Credibility qualification is distinct for other forms of qualification such as probabilistic and possibilistic (see Prade [1] for a comprehensive discussion of these kinds of qualifications). One manifestation of this distinction is that zero credibility reflects itself into a total lack of knowledge, rather than the negation or dual of the original knowledge as is the case in probability. Thus if an agent says that it is raining and if I assign no credibility to this agent's information rather then concluding it is not raining, I conclude nothing about the weather. In this regard, the operation of credibility qualification acts like an importance operator in multi-criteria decision making [2-6].

Another interesting characteristic of credibility qualification is that in some cases the credibility assigned to a piece of evidence can be a function of its compatibility with other, higher priority evidence. We call such credibility assignments relative credibility values. The structure of such credibility formulations can be seen to be very much in the spirit and form of nonmonotonic logic.[7] These types of structures play an important role in allowing credibility assignments to adjudicate and relieve conflicts between sources of information by allowing us to withdraw less credible information when it directly conflicts with more reliable information. In such cases the credibility of a piece of information can be said to be context dependent.

We shall specifically look at the issue of credibility qualification in the framework of Zadeh's theory of approximate reasoning [8]. We shall consider the situation where credibility is context dependent. In this situation, we shall draw upon Yager's work on nonmonotonic reasoning [9, 10]. In a more comprehensive study [4] we look at the issue of credibility qualification in the Dempster-Shafer theory of evidence [11, 12].

## 2. Credibility and Discounting in Approximate Reasoning

In this section we shall investigate the issue of credibility and discounting in the theory of approximate reasoning. A detailed discussion of the theory of approximation reasoning (AR) can be found in [8].

Assume V is a variable taking its value in the set X. A proposition (piece of knowledge) in AR is a statement of the form

$$V \text{ is } A.$$

where A is a fuzzy subset of X. The semantics of the above statement is to express the knowledge that the value of V lies in the set A. More formally, as described by Zadeh [13], such a statement induces a possibility distribution on X such that for each $x \in X$, $A(x)$ indicates the possibility that V assumes the value x.

In AR if we have a collection of propositions $P_1, P_2, \ldots P_n$ each of the form V is $A_j$, we indicate that a proposition P is inferable from our knowledge base as

$$(P_1, P_2, \ldots P_n) \vdash P$$

Specifically we can infer any proposition of the form V is B where B is any set such that

where
$$B(x) = D(\alpha, A(x))$$
then it is always the case that
$$B(x) \geq A(x)$$
that is $A \subset B$.

The following theorem relates the inference process to the discounting process.

**Theorem :** Reducing the credibility of any of the propositions in a knowledge base decreases the collection of valid inferences.

While a reduction of credibility has the effect of diminishing the number of valid inferences on one hand on the other hand it can play a significant role in reducing conflict between propositions in the knowledge base. As we shall subsequently see when we discuss the concept of relative credibility the use of credibility discounting can allow us to adjudicate and subsequently eliminate conflicting propositions in a systematic manner.

We have assumed that the credibility associated with a proposition was a precise number in the unit interval. In some cases our knowledge about the credibility associated with a proposition may not be good enough to allow us to so precisely indicate the value of $\alpha$. In these cases, we may find it useful to use a linguistic value to indicate the credibility of a proposition.

Consider the proposition
$$V \text{ is } A \text{ is } \alpha \text{ credible}$$
where $\alpha$ is a linguistic value such as "high", "low", "about 6". In this environment, we can represent $\alpha$ as a fuzzy subset of the unit interval. In this case for each $y \in [0,1]$, $\alpha(y)$ indicates the degree to which $y$ satisfies the concept, linguistic value, conveyed by $\alpha$. Some examples of notable linguistic values are shown in the figure 1.

In this environment, with $\alpha$ a linguistic value, the proposition
$$V \text{ is } A \text{ is } \alpha \text{ credible}$$
again induces an unqualified statement
$$V \text{ is } B$$
where
$$B(x) = D(\alpha, A(x)) = S(\bar{\alpha}, A(x))$$
In this situation $\bar{\alpha}$ is defined such that
$$\bar{\alpha}(y) = \alpha(1 - y).$$
In this case $\bar{\alpha}$ is the antonym of $\alpha$. This definition preserves the fact that
$$\bar{\alpha} = 1 - \alpha.$$
In this, since $\alpha$ is a fuzzy subset, it turns out that $B(x)$ becomes a fuzzy membership grade. That is, $B$ becomes a type II fuzzy subset, one with fuzzy membership grade.

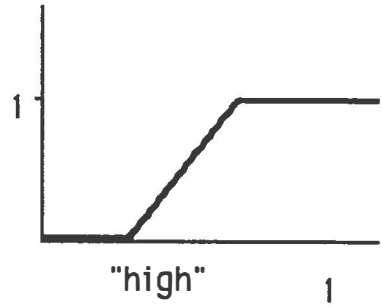
"high"

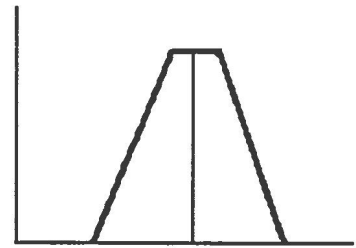
"close to 5"

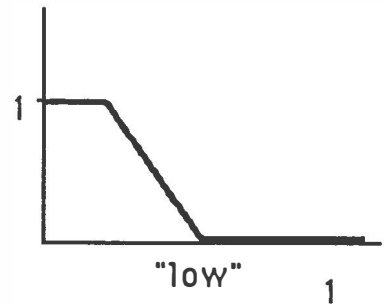
"low"

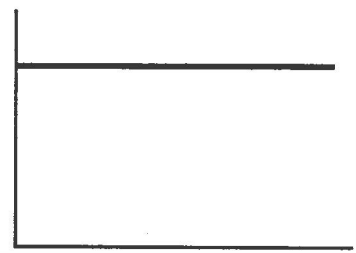
"unknown"

**Figure #1**

Consider the situation where we use for S the max operator, then
$$B(x) = \bar{\alpha} \vee A(x) = \alpha$$
In this case

$$B(x) = \cup_y \left\{ \frac{1 - \alpha(y)}{y \vee A(x)} \right\}.$$

If we used
$$D(x) = a^\alpha$$
then we would get
$$B(x) = \cup_y \left\{ \frac{\alpha(y)}{(A(x))^y} \right\}.$$

The following example will illustrate the calculation of B from A and $\alpha$ in the case where $\alpha$ is a fuzzy value.

**Example** : Let $X = \{a, b, c\}$. Assume we have the proposition
$$V \text{ is } A \text{ is } \alpha \text{ credible}.$$
Let
$$A = \left\{ \frac{.6}{a}, \frac{1}{b}, \frac{.8}{c} \right\}$$
and assume $\alpha$ has the value "low" where low is defined by
$$\text{"low"} = \{\frac{1}{0}, \frac{1}{.1}, \frac{.9}{.2}, \frac{.5}{.3}, \frac{.2}{.4}\}.$$
This proposition then induces a new proposition
$$V \text{ is } B$$
where B is a second order fuzzy set. Using the max type discounting function we get
$$B(x) = (\bar{\alpha} \vee A(x))$$
In this example
$$\bar{\alpha} = \{\frac{.2}{.6}, \frac{.5}{.7}, \frac{.9}{.8}, \frac{1}{.9}, \frac{1}{1}\}$$
Introducing this value into B(x) we get
$$B(a) = \{\frac{.2}{.6}, \frac{.5}{.7}, \frac{.9}{.8}, \frac{1}{.9}, \frac{1}{1}\} \vee \{\frac{1}{.6}\} = \{\frac{.2}{.6}, \frac{.5}{.7}, \frac{.9}{.8}, \frac{1}{.9}, \frac{1}{1}\}$$
$$B(b) = \{\frac{.2}{.6}, \frac{.5}{.7}, \frac{.9}{.8}, \frac{1}{.9}, \frac{1}{1}\} \vee \{\frac{1}{1}\} = \{\frac{1}{1}\}$$
$$B(c) = \{\frac{.2}{.6}, \frac{.5}{.7}, \frac{.9}{.8}, \frac{1}{.9}, \frac{1}{1}\} \vee \{\frac{1}{.8}\} = \{\frac{.9}{.8}, \frac{1}{.9}, \frac{1}{1}\}$$

## 3. Relative Credibility

In the previous section we associated with each proposition a value $\alpha \in [0,1]$ indicating the credibility we assigned to that piece of knowledge. The $\alpha$ value played a central role in determining how much we discounted the information provided in that proposition. This value was used simply as a measure of the credibility of the agent supplying the information. In this section, we shall consider the credibility assigned to a piece of evidence as determined by the compatibility of the piece of evidence with a collection of other pieces of evidence.

Assume we have a knowledge base consisting of $P_1, P_2, \ldots P_n$. Let

P: V is E

be a piece of evidence whose credibility $\alpha$ is denoted
$$\alpha = C(P_1, P_2, \ldots P_n).$$
In this framework the $P_1, \ldots P_n$ are called the **preeminent** knowledge associated with this piece of evidence. This formulation is meant to indicate that the discounting associated with the proposition V is E, its $\alpha$ value, is determined by the compatibility of the proposition, V is E, with the preeminent knowledge.

In order to obtain the compatibility of our proposition with its preeminent knowledge, we use the measure of possibility [13] of the knowledge given the preeminent knowledge
$$C_p(P_1, P_2, \ldots P_n) = \text{Poss}[P/P_1 \cap P_2 \cap \ldots P_n]$$
Assuming each $P_j$ is of the form
$$V \text{ is } A_j$$
then
$$C_p(P_1, P_2, \ldots P_n) = \text{Max}_x [A_1(x) \wedge A_2(x) \ldots \wedge A_n(x) \wedge E(x)].$$
If we denote
$$A = \cap_j A_j$$
then
$$\alpha = C_p(P_1, P_2, \ldots P_n) = \text{Poss}[V \text{ is } E / V \text{ is } A] = \text{Max}_x [A(x) \wedge E(x)].$$

Thus we see that the larger the degree of intersection between the knowledge of concern and its preeminent knowledge, the less we discount, the larger $\alpha$ value associated with the knowledge P. It should be carefully noted that the process of calculating $\alpha$ involves the whole of the sets P and A. This process is very much in the spirit of non-monotonic logics. [7].

Having obtained this $\alpha$ value then the piece of knowledge, V is E, gets discounted, and translated into
$$V \text{ is } F$$
where as before
$$F(x) = S(\bar{\alpha}, E(x)).$$
However since in this case
$$\alpha = \text{Poss}[E/A]$$
we get
$$F(x) = S(1 - \text{Poss}[E/A], E(x)).$$
Furthermore if we use the max operator for S than
$$F(x) = E(x) \vee (1 - \text{Poss}(E/A))$$
If we use the exponential model then
$$F(x) = E(x)^{\text{Poss}[E/A]}.$$

We note using this general formalism that if Poss[E/A] = 1 than no discounting takes place. On the other hand if Poss[E/A] = 0 complete discounting has occurred.

In a more general sense, we can express the relationship between $\alpha$ and Poss[E/A] in a functional manner, ie

$$\alpha = g(Poss(E/A))$$

where g is monotonically increasing function. If $g(1) < 1$ then the evidence is discounted even if it is completely compatible with the preeminent knowledge

We first note that if for all x

$$g(x) = a$$

then we are essentially in the situation of the previous section, one in which a is the absolute measure of credibility and is not effected by the preeminent knowledge. In the special case where

$a = 1$ then the information is not discounted at all.

## 4. Alternative Interpretations of Credibility

In the previous sections, when faced with a proposition whose credibility was less than complete we discounted the information provided by the proposition essentially by making less specific the information provided. Thus given

V is A is $\alpha$ credible

we derived the information

V is $A^+$

where

$$A^+(x) \geq A(x) \text{ for } x.$$

Fundamentally, we ended up with another, although less informative, proposition of the same type. One way of viewing the effect of this operation is in terms of possibility and certainty measures[15]. In particular we recall that for any subset F

$$Poss[V \text{ is } F / V \text{ is } E] = Max_x[F(x) \wedge E(x)]$$
$$Cert[V \text{ is } F / V \text{ is } E] = 1 - Poss[\overline{F}/E]$$

Since $A^*(x) \geq A(x)$ then for all subsets B

$$Poss[V \text{ is } B/ V \text{ is } A^+] \geq Poss[V \text{ is } B / V \text{ is } A]$$

$$Cert[V \text{ is } B / V \text{ is } A^+] \leq Cert[V \text{ is } B / V \text{ is } A]$$

Therefore we essentially increased what is possible and decreased what is certain.

There appears to be an alternative way of discounting a piece of evidence to reflect our lack of complete credibility in the information. This approach can consist of a probabilistic type of discounting. Assume we have a piece of evidence

V is A

which has a credibility $\alpha$. An alternative way of discounting is to assign a probability $\alpha$ that V is A is indeed correct and a probability $1 - \alpha$ to the statement

V is X.

This type of operation generates a Dempster-Shafer belief structure, a simple support function [12, 16]:

V is A $\alpha$

V is X $1 - \alpha$

We note that in this case if our credibility is zero, we end up with the complete base set

V is X,

while if $\alpha = 1$, we end up with the original piece of evidence, V is A.

We recall that in the Dempster-Shafer framework for any subset B of X, we can define two set measures plausibility (Pl) and belief (Bel) such that for any set B

$$Pl(B) = \alpha Poss[B/A] + (1 - \alpha)Poss[B/X]$$
$$Bel[B] = \alpha Cert[B/A] + 1 - \alpha \ Cert[B/X]$$

where

$$Cert(B) \leq Prob(B) \leq Pl(B).$$

In the following, we shall assume A is our known proposition which has credibility $\alpha$ and B is some arbitrary proposition of interest. We shall assume that both A and B are normal, that is

$$Max_x A(x) = 1 \text{ and } Max_x B(x) = 1.$$

We first note that if $\alpha = 1$ then

$$Pl(B) = Poss[B/A] \text{ and } Bel(B) = Cert[B/A].$$

Thus plausibility and possibility are effectively measuring the same concept while certainty and belief are measuring the same concept. On the other hand if $\alpha = 0$ then

$$Pl(B) = Poss[B/X] = 1$$
$$Cert(B) = Cert[\overline{B}/X] = Min_x B(x).$$

Let us denote

$$Pl_\alpha(B) = \alpha Poss[B/A] + (1 - \alpha) Poss[B/X]$$
$$Bel_\alpha(B) = \alpha Cert[B/A] + (1 - \alpha) Cert[B/X]$$

Since

$$Poss[B/X] = 1$$

then

$$Pl_\alpha(B) = \alpha Poss[B/A] + 1 - \alpha.$$

In the special case when $\alpha = 1$, we shall drop the subscript.

The following relationships hold for all A, B and $\alpha$.

$$Pl(B) \leq Pl_\alpha(B)$$

since

$$Poss[B/A] \leq \alpha Poss[B/A].$$

Furthermore

$$Bel[B] \geq Bel_\alpha[B]$$

since

$$Cert[B/X] \leq Cert[B/A].$$

These relationships indicate that the introduction of discounting via this probabilistic method also results in an increase in uncertainty.

The belief structure resulting from this discounting is a consonant belief structure [12] since $A \subset X$. As shown in Shafer [12] consonant belief structures have some very unique properties. In particular, for any consonant belief structure there exists a contour function $F(x)$ such that

$$F(x) = Pl(\{x\}).$$

In our case

$$F(x) = \alpha A(x) + 1 - \alpha$$

The contour function has the characteristic that

$$Pl_\alpha(B) = Max_X[F(x) \wedge B(x)]$$

and since

$$Bel(B) = 1 - Pl(\bar{B})$$

it follows that

$$Bel_\alpha(B) = 1 - Max_X[F(x) \wedge \bar{B}(x)].$$

In our case this implies

$$Pl_\alpha(B) = Max_X[(\alpha A(x) + (1-\alpha)) \wedge B(x)]$$
$$Bel_\alpha(B) = 1 - Max_X[(\alpha A(x) + (1-\alpha)) \wedge \bar{B}(x)]$$

We next recall from the previous section that our methodology for discounting a piece of evidence

V is A

which had credibility $\alpha$ was to form a new piece of evidence

V is D

such that each $x \in X$

$$D(x) = S(\bar{\alpha}, A(x))$$

where S is any arbitrary co t-norm. Let us consider the special t-conorm

$$S(a,b) = a + b - ab$$

Using this operator

$$D(x) = \bar{\alpha} + A(x) - \bar{\alpha} A(x)$$
$$= \alpha A(x) + (1 - \alpha)$$

From this we obtain

$$Poss[B/D] = Max_X[(\alpha A(x) + (1 - \alpha)) \wedge B(x)]$$
$$Cert[B/D] = 1 - Max_X[(\alpha A(x) + (1 - \alpha)) \wedge \bar{B}(x)]$$

From this we easily see that

$$Poss[B/D] = Pl_\alpha(B)$$
$$Cert[B/D] = Bel_\alpha(B)$$

Thus we see that the use of probabilistic type discounting can be seen as a special case of our original form of discounting with the appropriate selection of the operator S.

There are a number of other factors we might want to consider in the formulation of credibility qualification. Assume we have a piece of evidence

V is A

obtained from some agent. It is conceivable that the agents credibility may be a function of the value x of the base set X underlying V. That is, a particular source maybe very effective in analyzing information in some range of X but not as good in other ranges. This observation leads us to consider the case in which the credibility assigned a piece of evidence is a function of x, that is $\alpha = h(x)$. In this situation

$$D(x) = S(1 - h(x), A(x)).$$

There exists another consideration we may desire to include in our transformation. Assume V is some variable which takes its value in the base set X. Furthermore assume that X has some metric on it which enables us to measure a degree of proximity between any two points. Let

$$p: X \times X \rightarrow [0,1]$$

where the larger the value $p(x,y)$ the closer the two elements. Assume we have a piece of evidence

V is A

where $A = \{x_0\}$. That is, the evidence states that V is $x_0$. Assume we have some credibility $\alpha$ assigned to this statement. Then

$$D(x) = S(1 - \alpha, A(x)).$$

Furthermore, in this case we would get

$$D(x_0) = 1$$
$$D(x) = 1 - \alpha \quad \text{for all } x = x_0.$$

Thus all the values of x not equal to $x_0$ have the same value for $D(x)$. In many situations this new D doesn't capture what we desire to happen. For example what may be a more appropriate formulation would be that D is around $x_0$. That is we would like to "soften the edges" at $x_0$ rather then make everything the same. (See figure #2)

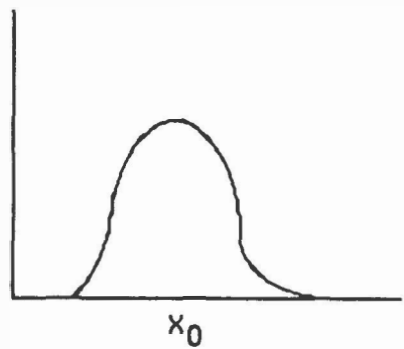

**Figure #2**

Thus in this case, we would like the discounting at a point x to be a function of the proximity of x to $x_0$. In this situation, we could have a discounting factor h such that

$$h(x) = f(\alpha, p(x, x_0)).$$

The form of f will determine the shape of D resulting from the discounting.

More generally, if A is a fuzzy subset we would use the proximately of x to the subset, that is p(x,A). One requirement on f is that if $\alpha = 0$, $f(\alpha,b)=0$ for all b and if $\alpha = 1$ then $f(\alpha,b) = 1$ for all b. It would also appear that continuity would be a desirable requirement. Furthermore monotonicity on F is required in that if $\alpha_1 > \alpha_2$ then

$$f(\alpha_1, b) \geq f(\alpha_2, b)$$

and if $b_1 > b_2$ then

$$f(\alpha, b_1) \geq f(\alpha, b_2)$$

## 5. Conclusion

We have introduced a type of qualification which we called credibility qualification of a piece of evidence. We have used its close relationship to the idea of importance to provide a mechanism for formally manipulating these qualifications. We indicated how this type of qualification essentially results in a discounting of the given evidence in a manner inversely related to its credibility. Significantly we indicated how the credibility associated with a piece of evidence can be obtained as a function of the compatibility of the given evidence with other pieces of more established evidence. This idea of relative credibility is very much in the spirit of nonmonotonic and default reasoning.